\documentstyle[11pt,epsf,named,twocolumn,dina4]{article}
\begin{document}


\title{Generating Multilingual Documents from a Knowledge Base: The TECHDOC Project }

\author{Dietmar R\"osner \\ FAW Ulm \\ P.O.\ Box 2060, 89010 Ulm, Germany\\ 
{\tt roesner@faw.uni-ulm.de}
\and
Manfred Stede \\ (University of Toronto and) FAW Ulm  \\ P.O.\ Box 2060, 
89010 Ulm, Germany\\ {\tt stede@faw.uni-ulm.de} }
\date{}
\maketitle

\thispagestyle{empty}

\begin{abstract}
TECHDOC is an implemented system demonstrating the feasibility of generating 
multilingual technical documents on the basis of a language-independent 
knowledge base. Its application domain is user and maintenance instructions, 
which are produced from underlying plan structures representing the 
activities, the participating objects with their properties, relations, and 
so on.  This paper gives a brief outline of the system architecture and
discusses some recent developments in the project: the addition of actual
event simulation in the KB, steps towards a document authoring tool, and
a multimodal user interface.
\end{abstract}

\section{Overview}

\subsection{Project idea}

The availability of technical documents in multiple languages is a
problem of increasing significance. Not only do consumers demand
adequate documentation in their mother tongue; there are also legal
requirements, e.g., with respect to the upcoming European common
market: the product reliability act forces merchants to offer complete
technical documentation in the consumer's native language.  The need to
provide such a massive amount of multilingual material is likely to
exceed both the capacities of human translators as well as those of
machine translation technology currently available.  Our work in the
TECHDOC project is motivated by the feeling that this situation calls
for investigating a potential alternative: {\em to exploit
natural language generation technology in order to help overcome the
documentation problem}.

TECHDOC operates in the domain of {\em technical manuals}, which was
selected for two principal reasons.  On the one hand, they represent
``real-world'' texts that are actually useful: the domain is {\em
practical} instead of a ``toy world''.  On the other hand, the
language that is used in such manuals tends to be relatively simple;
one mostly finds straightforward instructions that have been written
with the intention to produce text that can be readily understood by a
person who is executing some maintenance activity.  Moreover, as our
initial analyses in the first phase of TECHDOC had shown, the {\em
structure} of manual sections is largely uniform and amenable to
formalization.

\subsection{Outline of the generation process}

TECHDOC produces maintenance instructions in English, German and French.  The
system is based on a KB encoding technical domain knowledge as well as
schematic text structure in LOOM, a KL-ONE dialect \cite{loom91}.  The {\em
macrostructure} of a manual section is captured by schemas saying that
(if appropriate) one first talks about the location of the object to be
repaired/maintained, then about possible replacement parts/substances;
next, the activities are described, which fall into the three general
categories of checking some attribute (e.g., a fluid level), adding a
substance and replacing a part/substance. These actions are represented
as plans in the traditional AI sense, i.e.\ with pre- and
postconditions, and with recursive structure (steps can be elaborated
through complete refinement plans).

These representations are mapped onto a language-independent document
representation that also captures its {\em microstructure} by means of RST
relations \cite{m&t87} with a number of specific annotations (e.g., a proposition is to
be expressed as an instruction, giving rise to imperative mood).  This
document representation is successively transformed into a sequence of sentence plans
(together with formatting instructions in a selectable target format; 
SGML, \LaTeX, Zmacs and --- for screen output --- slightly formatted ASCII are
currently supported), which are handed over to
sentence generators.  For English, we use `Penman' and its sentence
planning language (SPL) as input terms.  To produce
German and French text, we have implemented a German version of Penman's grammar
(NIGEL), which is enhanced by a morphology module, and a fragment of a French grammar
in the same way.

For a more detailled description of the system architecture see 
\cite{konvens92}.
\thispagestyle{empty}

\section{The Knowledge Base}

The Knowledge Base is encoded in LOOM.
In addition to the standard KL-ONE functionality (structured
inheritance, separation of terminological and assertional knowledge),
LOOM supports object-oriented and also rule-based programming.  

In addition to the `Upper Model' of the Penman generator (a basic
ontology that reflects semantic distinctions made by language, \cite{bat90}) more
than 1000 concepts and instances constitute the TECHDOC KB.  They
encode the technical knowledge as well as the plan structures that
serve as input to the generation process. The domains currently
modeled are end consumer activities in car maintenance and some
technical procedures from an aircraft maintenance manual.

One of the central aims in the design philosophy of the TECHDOC 
knowledge base is the separation of domain-independent technical
knowledge and specific concepts pertaining to the particular domain: the
portability of general technical knowledge has been a concern from
the beginning.  For instance, knowledge about various types of {\em tanks}
(with or without imprinted scales, dipsticks, drain bolts) is encoded
on an abstract level in the inheritance network (the `middle model'), and the particular
tanks found in the engine domain are attached at the lower end.
Similarly, we have an abstract model of {\em connections} (plugs,
bolts, etc.), their properties, and the actions pertaining to them
(plug-in connections can be merely connected or disconnected, screw
connections can be tightly or loosely connected, or disconnected).
Objects with the functionality of connections (e.g., spark plugs)
appear at the bottom of the hierarchy.  Thus, when the system is
transferred to a different technical domain --- as experienced recently
when we moved to aircraft manuals ---, large parts of the abstract
representation levels are re-usable.

\section{Document Representation Using RST}

\thispagestyle{empty}

The first task undertaken in TECHDOC was a thorough analysis of 
a corpus of pages from multilingual manuals in terms of {\em content} as well as {\em structure} of the
sections.  A text representation level was sought that captured the
commonalities of the correponding sections of the German, English and French texts, i.e.\ that was not tailored
towards one of the specific languages (for a discussion of representation
levels in multilingual generation, see \cite{eacl93}).  Rhetorical Structure
Theory (RST) turned out to be a useful formalism: for almost every section
we investigated, the RST trees for the different language versions were
identical.  

Our work with RST gave rise to a number of new discourse relations that we
found useful in analyzing our texts.  Also, we discovered several general
problems with the theory, regarding the status of minimal units for the
analysis and the requirement that the text representation be a tree 
structure all the time (instead of a general graph).  These and other
experiences with RST are reported in \cite{r&snlgws92}.

\section{Recent Developments}

\subsection{Event simulation in the knowledge base}

We developed a detailled representation of knowledge about
actions. Together with an action concept, preconditions and
postconditions can be defined in a declarative way. The preconditions
can be checked against the current state of the knowledge base (via
LOOM's ASK queries). If the preconditions hold, the action can be
performed and the postconditions are communicated to the knowledge
base (with the TELL facility of LOOM).  This typically leads to
re-classification of certain technical objects involved. With the help
of LOOM's production rule mechanism, additional actions either in the
knowledge base or on an output medium (e.g., for visualization) can be
triggered.  In this mode, instruction generation is a by-product of 
simulating the actions that the instructions pertain to.

Being able to take the current state of a technical device into account,
as in this simulation mode, is a prerequisite for upcoming interactive
applications of instruction generation: devices equipped with adequate
sensory instruments produce raw data that can be fed directly into the 
knowledge base.  Thereby, the specific situation of the device, e.g., the
car, drives the instruction generation process, so that only the truly
relevant information is given to the user.

\subsection{Towards a document authoring tool}

A first version of an authoring tool has been designed and implemented
and tested with a number of users. The authoring tool allows to 
interactively build up knowledge base instances of maintenance plans,
including the actions and objects involved, and to convert them 
immediately into documents in the selected languages.  At any time,
the tool takes the current state of the knowledge base into account:
all menus offering selections dynamically construct their selection lists,
so that only options of applicable types are offered.

\subsection{From text generation to a multimodal information system}

The generated texts are now displayed with words, groups and phrases
and whole sentences being mouse-sensitive and --- when selected ---
offering menus with applicable queries to be directed to the
underlying knowledge base instances.  This allows for a number of
tasks to be performed on the generated surface texts, for example:

\begin{itemize}
\item pronouns can be asked about their antecedent referent,

\item linguistic items in the output for one language can be asked about
their corresponding items in the other languages output,

\item objects can be asked about their location, answered by a suitable 
graphic illustration,

\item actions can be asked for more detailled instructions on how to 
perform them, answered by a short video sequence.
\end{itemize}

In essence, these facilities have paved the way to move from static,
inactive strings as output to an active and dynamic interface for the
associated knowledge sources and their various presentation
modalities.  The key is that all information types (lexemes in various
languages, images and object's location therein, and video sequences)
are associated with the underlying KB instances, which are in turn
linked to their referents in the mouse-sensitive output text.  Figure
1 shows a sample screen, where the user has just asked for
additional ``location'' information about the dipstick, by clicking on
the word in one of the text output windows.

\thispagestyle{empty}

\begin{figure*}[top]
\epsffile{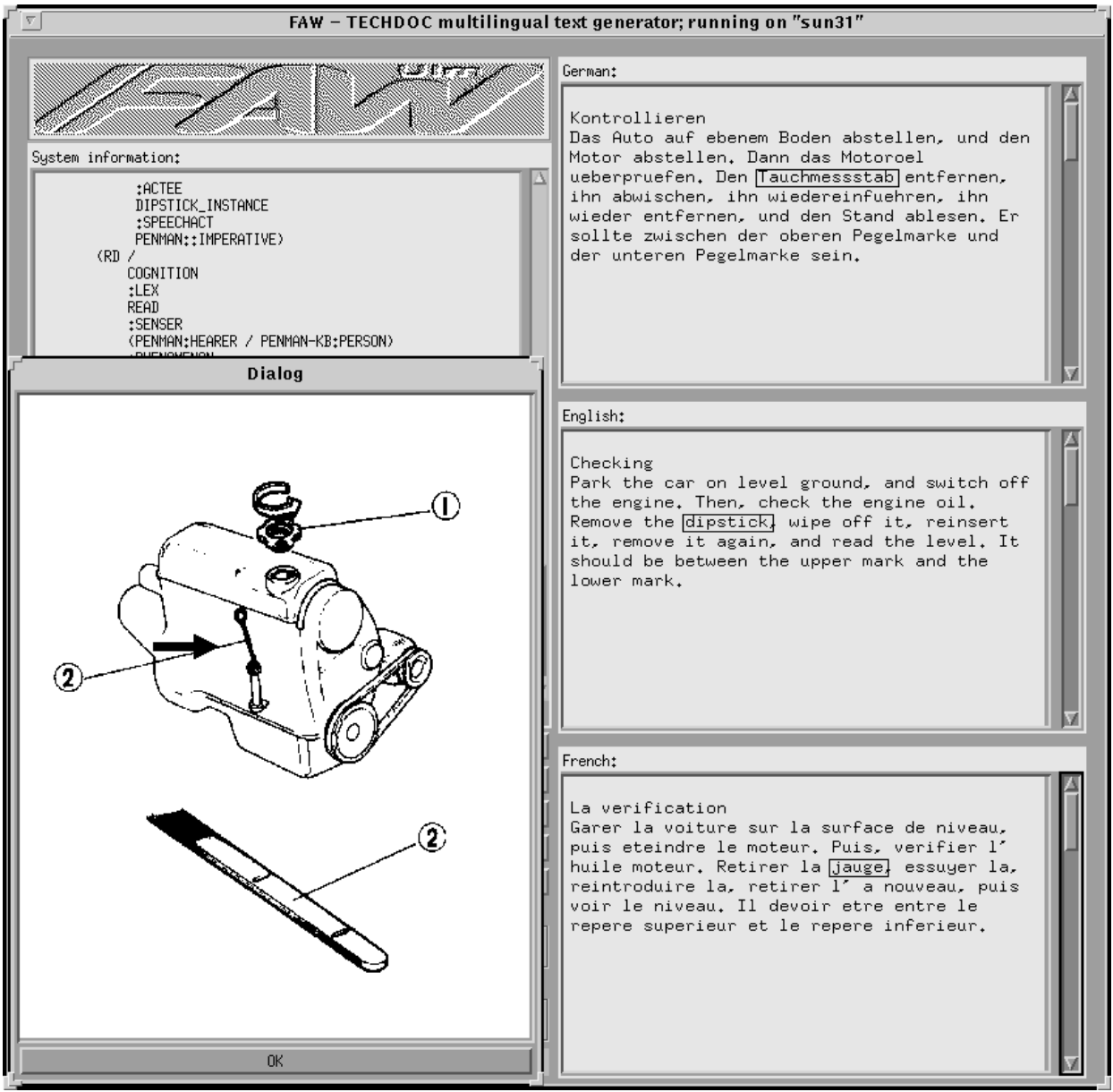}
\caption{Trilingual output and interactive graphic support}
\end{figure*}


\section{Implementation Note}

The current version of TECHDOC is running on Sun Sparc stations with
LUCID CommonLISP 1.4 and LOOM 1.41 (a port to LOOM 2.1 is underway), 
and a PENMAN version from 1991.
The user interface is based on the CommonLISP Motif interface
package CLM and the application building tool GINA \cite{spenke2-92}.

\footnotesize
\section*{Acknowledgements}

\noindent The success of the TECHDOC project depended heavily on 
contributions from a number of student interns, in alphabetical order:
Brigitte Grote, Sandra K\"ubler, Haihua Pan, Jochen Schoepp, Alexander 
Sigel, Ralf Wagner, and Uta Weis. They all have contributed to grammar 
or lexicon coverage in one way or another. Gerhard Peter has implemented
TECHDOC-I, an
interactive version giving car maintainance assistance. 
Thorsten Liebig has implemented TECHDOC's user interface for
workstations using CLM and GINA, Hartmut Feuchtm\"uller has added
multimedia facilities and mouse-sensitive text output. We also have
to thank the PENMAN and LOOM groups at USC/ISI and the KOMET project at GMD
Darmstadt, who gave us invaluable help.


\bibliographystyle{/users/stede/Latex/named}

\bibliography{/users/techdoc/tex/lit/gengeneral}

\begin{thebibliography}{}

\bibitem[\protect\citeauthoryear{Bateman}{1990}]{bat90}
John~A. Bateman.
\newblock Upper modeling: A level of semantics for natural language processing.
\newblock In {\em Proceedings of the Fifth International Workshop on Natural
  Language Generation}, Pittsburgh, PA., 3 - 6 June 1990.

\bibitem[\protect\citeauthoryear{Grote \bgroup \em et al.\egroup
  }{1993}]{eacl93}
Brigitte Grote, Dietmar R\"osner, and Manfred Stede.
\newblock Representation levels in multilingual text generation.
\newblock In Brigitte Grote, Dietmar R\"osner, Manfred Stede, and Uta Weis,
  editors, {\em From Knowledge to Language -- Three Papers on Multilingual text
  Generation}. FAW Ulm, FAW-TR-93017, 1993.

\bibitem[\protect\citeauthoryear{LOOM}{1991}]{loom91}
The {LOOM} {K}nowledge {R}epresentation {S}ystem.
\newblock Documentation package, USC/Information Sciences Institute, Marina Del
  Rey, CA., 1991.

\bibitem[\protect\citeauthoryear{Mann and Thompson}{1987}]{m&t87}
William~C. Mann and Sandra~A. Thompson.
\newblock Rhetorical structure theory: A theory of text organization.
\newblock In L.Polanyi, editor, {\em The Structure of Discourse}. Ablex,
  Norwood, N.J., 1987.
\newblock Also as USC/Information Sciences Institute Research Report RS-87-190.

\bibitem[\protect\citeauthoryear{R\"osner and Stede}{1992a}]{r&snlgws92}
Dietmar R\"osner and Manfred Stede.
\newblock Customizing {RST} for the automatic production of technical manuals.
\newblock In R.~Dale, E.~Hovy, D.~R\"osner, and O.~Stock, editors, {\em Aspects
  of Automated Natural Language Generation - Proceedings of the 6th
  International WS on Natural Language Generation}, Lecture Notes in Artificial
  Intelligence 587. Springer, Berlin/Heidelberg, 1992.

\bibitem[\protect\citeauthoryear{R\"osner and Stede}{1992b}]{konvens92}
Dietmar R\"osner and Manfred Stede.
\newblock {TECHDOC} : A system for the automatic production of multilingual
  technical documents.
\newblock In G.~G\"orz, editor, {\em KONVENS 92}, Reihe Informatik aktuell.
  Springer, Berlin/Heidelberg, 1992.

\bibitem[\protect\citeauthoryear{Spenke \bgroup \em et al.\egroup
  }{1992}]{spenke2-92}
Michael Spenke, Christian Beilken, Thomas Berlage, Andreas B"acker, and Andreas
  Grau.
\newblock {\em GINA Reference Manual Version 2.1}.
\newblock German National Research Center for Computer Science, Sankt Augustin,
  Germany, 1992.

\end{thebibliography}

\footnotesize
\section*{Bibliographical note}

This is a slightly updated version of a paper published in:

{\bf COLING-94, Proceeedings, Kyoto 1994}

\end{document}